\documentclass{llncs}
\usepackage{times}
\usepackage{amssymb}
\usepackage{amstext}
\usepackage{amsmath}
\usepackage{eucal}
\usepackage{graphics}
\usepackage{graphicx}
\usepackage[T1]{fontenc}
\usepackage{ae,aecompl}
\usepackage{zeta}
\usepackage{graphs}
\usepackage{tikz}
\usepackage{multicol}
\usepackage{algorithm}

\begin{document}

\pagestyle{empty}

\title{Justificatory and Explanatory Argumentation for Committing Agents}
\author{Ioan Alfred Letia\inst{1}
\and Adrian Groza\inst{1}}
\institute{Technical University of Cluj-Napoca\\
Department of Computer Science\\
Baritiu 28, RO-400391 Cluj-Napoca, Romania\\
\email{\{letia,adrian\}@cs-gw.utcluj.ro}\\
}

\maketitle

\thispagestyle{empty}

\begin{abstract}
In the interaction between agents we can have an explicative discourse, 
when communicating preferences or intentions, and a normative discourse, when considering 
normative knowledge. For justifying their actions
our agents are endowed with a Justification and Explanation Logic ($\mathcal{JEL}$),
capable to cover both the justification for their commitments and 
explanations why they had to act in that way, due to the current situation
in the environment. 
Social commitments are used to formalise justificatory and explanatory patterns. 
The combination of explanation, justification, and commitments provides flexibility for defining several types of argumentative agents.
\end{abstract}

\section{Introduction}

The institutional economics demands a clear account of a fuzzy world
and is therefore in accordance~\cite{Morgan11JIE} with
the dictum attributed to Keynes:
"it is better to be roughly right than precisely wrong".
As many multi-agent systems are organized in some kind of institution,
judging the behavior of agents in such contexts requires sometimes understanding
of how their actions stand vis-a-vis to the goals of the overall system.

Habermas~\cite{Morgan11JIE} proposed different types of discourses dealing
with different validity claims. In the interaction between agents we can thus
have an explicative discourse, when communicating knowledge, and a normative
discourse, when considering normative knowledge. For justifying their actions
our agents are endowed with a Justification and Explanation Logic ($\mathcal{JEL}$),
capable to cover both the justification for their commitments and 
explanations why they had to act in that way, due to the current situation
in the environment. The distinction between factual and normative knowledge
is important in argumentative agents since facts of various kinds might not
provide agents with any choice, while norms are possible to be bent and
eventually changed.

The difference between explanation and justification has not been clearly
delimited in computational models of arguments.
From this perspective, there is a gap between argumentation in the philosophy
of science and computational-based argumentation.
Given the idea that argumentation is a means to justify claims or to
persuade other agents of these claims~\cite{Jennings98}, there are two
approaches when defining what "good argumentation" represents:
i) argumentation able to justify its target claims and 
ii) argumentation able to convince an audience.
These two lines form the basis for investigating the contrast between
\textit{justificatory arguments} and \textit{explanatory arguments}.

The philosophy of science distinguishes between \textit{explanatory
reasons} and normative or \textit{justificatory reasons}~\cite{sep-reasons-just-vs-expl}.
While explanations are reasons why events occur, justifying reasons are
"considerations which count in favor" or "explanations of ought facts".
Differently, in the computation models of arguments there is a blurred
distinction between explanation-based argumentation and
justification-based argumentation, as in~\cite{Lu2011946}, where
explanations and evidence are used to construct justifications.

In many practical domains the distinction between explanation and
justification has pragmatic force.
Many examples come from the legal domain, where an illegal action can be
explained, but in many cases the action is not justified.
For instance, a theft can be explained by the loss of money because the
person lost his job, but it does not have normative justificatory power.
Another common example is based on the well known lack of time explanatory
pattern: "I could not finalize the task due to lack of time", with
its many instantiations: "I could not review the paper because I was at
a conference during the deadline" or "I could not finalize the article,
because input data arrived too late".
In most situations, this explanation pattern does not have
justificatory power at all.
In other words, something can explain a behavior but it cannot always
justify it.

In this paper we investigate the inter-living of explanations and
justifications in business-oriented situations.
Current business interactions are affected by postmodern ideas like
post-structuralism and heterogeneity at different levels.
In line with postructuralism, business commitments act as a flexible
framework for guiding business interactions, agents preferring soft law against
the hard law governance, 
whilst the idea of heterogeneity of individuals is reflected at the
business policy level through the concept of heterogeneity of
customers~\cite{Yan10}.
Consider the contractual clause in which the debtor $a$ promises to
deliver an item to the creditor $b$ within a pre-agreed deadline.
Not meeting the deadline, the agent $a$ provides an explanation: 
"I could not deliver the product because my supplier $c$ has not delivered the
parts yet".
This explanation provides the creditor $b$ with some insights on the
current situation, but it does not have enough justificatory power in
order to justify the behavior of agent $a$ for not delivering the item.
It may be a case in which the supplier $c$ has no normative obligation
to deliver the components of the product: either a commitment 
$C(s, a, pay, parts)$ for delivering the parts does not exist between $a$ and 
the supplier $c$, or the agent $a$ did not pay the components in due time.
Valid justifications should be normative, like: 
"emergency situation forces me not to deliver the item",
where emergency is a normative concept.
Observe that the $emergency$ justifier does not help the debtor $a$ to
understand the situation.
The agent can ask for further explanations for the conveyed justifier, but
also for further normative justifications for it.

The paper advances the state-of-the-art in logic of argumentation in three ways: 
i) proposing $\mathcal{JEL}$ for handling both justificatory and explanatory arguments; 
ii) introducing commitments as proof terms; and 
iii) formalising several justificatory and explanatory patterns. 
The remaining of the paper is organised as follows:
Section~\ref{sec:logic} extends justification logic~\cite{Artemov08} with explanatory capabilities and introduces commitments as proof terms. 
The expressivity of social commitments is exploited in section~\ref{sec:patterns} to represent both justificatory and explanatory patterns. 
Section~\ref{sec:agents} describes types of argumentative agents based on the combination of justificatory and explanatory attitudes, whilst section~\ref{sec:scenario} illustrates the developed instrumentation through an illustrative scenario.  
We end the paper with related work and conclusions.

\section{Commitment-Based Justification and Explanation}
\label{sec:logic}

The first part of this section extends the justification logic with explanatory capabilities. 
The second part illustrates how social commitments are used to represent proof terms in justification logic.  

\subsection{Enhancing Justification Logic with Explanation}


In order to model justificatory and explanatory arguments we propose an
extended version of the Justification Logic (JL).
Justification Logic combines ideas from epistemology and the mathematical
theory of proofs.
It provides an evidence-based foundation for the logic of knowledge,
according to which "F is known" is replaced by "F has an adequate
justification".

Simply, instead of "X is known" ($KX$) consider $t:X$, that is, "X is
known for the explicit reason t"~\cite{Artemov08}.
Justification logic lacks this component of interpretation.
It also lacks the capability to express explanation or partial justification.
This section extends the justification logic with explanatory
capabilities, by introducing the explanatory operator $t \triangleleft F$,
where $t$ is an explanation for $F$.

\begin{definition}
The language of Justification and Explanation Logic $\mathcal{JEL}$ contains proof terms $t\in \mathcal{T}$ and
formulas $F \in \mathcal{F}$
\begin{center}
\begin{tabular}{ll}
$t:$& $=x |c |t\cdot t | t + t | !t| ?t | t \gtrdot t$\\
$F:$ &$=p| F \vee F |\neg F | t: F | t \triangleleft F$
\end{tabular}
\end{center}
\end{definition}

Proof terms $t$ are abstract objects that have structure.
They are built up from axiom constants $c\in Cons$, proof variables $x\in
Vars$, and operators on justifications and explanations
$\cdot$, $+$, $!$,$?$,$\gtrdot$. 
The application operator $\cdot$ takes two proof terms and constructs a new justification based on them. 
The sum operator $+$ concatenates the given proofs, whilst the unary operators $!$ and $?$ are used to request for positive and negative proof terms for a formula in $\mathcal{JEL}$.  
The operator precedence decreases as follows: $!, \cdot, +, :,\triangleleft, \neg, \vee$, and
 $\cdot, +$ are left associative, and $:,\triangleleft$ right associative.
To express that $t$ is not probative justification for supporting $F$ one
uses $\neg t:F$.
Parentheses are needed to express that $\neg t$ is a justification for
$F$: $(\neg t):F$.
Note that justification is used to support negated sentences too, as in
$t:\neg F$.
Similar semantics applies for the explanation operator $\triangleleft$.

\begin{figure}
\begin{tabular}{lllr}
\hspace{0cm}$A_0$ & classical propositional&  axioms &\\
\hspace{0cm}$A_1$ & $F \rightarrow$ & $(t: F \vee t\triangleleft F)$ &
(necessity)\\ 
\hspace{0cm}$A_2$ & $s: (F\rightarrow G) \rightarrow $ & $(t: F\rightarrow (s \cdot t): G)$ & (j-application)\\
\hspace{0cm}$A'_2$ & $s\triangleleft (F\rightarrow G) \rightarrow $ & $
(t\triangleleft F\rightarrow (s \cdot t)\triangleleft G)$ & (e-application)\\
\hspace{0cm}$A''_2$ & $s: (F\rightarrow G) \rightarrow $ & $(t\triangleleft F\rightarrow (s \cdot t)\triangleleft G)$ & (e-application)\\
\hspace{0cm}$A'''_2$ & $s\triangleleft (F\rightarrow G) \rightarrow $ & $(t: F\rightarrow (s \cdot t)\triangleleft G)$ & (e-application)\\
\hspace{0cm}$A_3$ & $s: F \rightarrow $ & $(s + t): F$ & (j-sum)\\
\hspace{0cm}$A_4$ & $t: F \rightarrow $ & $ !t:(t: F) \vee !t\triangleleft(t: F)$& (proof checker)\\
\hspace{0cm}$A'_4$ & $t\triangleleft F \rightarrow $ & $!t\triangleleft(t\triangleleft F)$ & (explanation checker)\\
\hspace{0cm}$A_5$ & $\neg t: F \rightarrow $ & $ ?t:(\neg t: F) \vee ?t\triangleleft (\neg t: F)$ & (negative proof checker)\\
\hspace{0cm}$A'_5$ & $\neg t\triangleleft F \rightarrow $ & $ ? t\triangleleft(\neg t\triangleleft F)$ & (negative explanation checker)\\
\end{tabular}
\caption{Axioms of $\mathcal{JEL}$.}
\label{fig:axioms}
\end{figure}

The axioms of $\mathcal{JEL}$ are shown in figure~\ref{fig:axioms}, where
axiom $A_1$ forces all formulas $F$ to have a justification or an explanation
$t$.
The compounds $t:F$ or $t\triangleleft F$ represent a formula, which
should have their own justification.
This corresponds to the \textit{principle of inferential justification}:
for sentence $F$ to be justified on the basis of $t$ one must justify $t$
and justify that $t$ makes $F$ plausible.
Constants are used to stop the ad infinitum justification chain by
representing a kind of justification that does not depend on other
justifiers.

The application axiom $A_2$ takes a justifier $s$ of an implication $F
\rightarrow G$ and a justifier $t$ of its antecedent $F$, and produces
a justification $s \cdot t$ of the consequent $G$.
If at least one of the terms $s$, $t$ represents an explanation, the
formula $G$ is considered only explained, but not justified 
(axioms $A'_2$, $A''_2$, and $A'''_2$).

The j-sum axiom says that if a formula $F$ is justified by the justifier
$s$, than for a new justifier $t$, the formula is still justified.
Thus, justification reasoning is monotonic, new justification not defeating
the existing one.
The corresponding axiom for explanation is missing, leaving space for
contradictory explanations and non-monotonic explanatory reasoning.
The rationality behind this is that norms are considered static at a given moment and
assumed apriori known by the participants, whilst explanations are
constructed dynamically during a dialog.

Justifications and explanations are assumed to be verified.
Based on axiom $A_4$, a justifier $t$ for formula $F$ can be further justified
by the term $!t$, but it can also be explained.
The axiom $A'_4$ limits the possibility to justify an explanation, an
explanation can only be further explained.
The negative proof checker $A_5$ forces agents to provide justifications
or explanations why they are not able to justify a particular formula $F$.
Justifiers cannot be used to justify why the formula $F$ is not explained
by the explanandum $t$, as axiom $A'_5$ states that only explanans can be used.
A proof term can be stronger than another one, given by the operator $\gtrdot$.

\subsection{Commitment Based Proof Terms}

One of the issues regards what sort of things can be a justifier.
In a normative framework regulated only by social commitments, a justifier
can be represented by such commitments.
In the proposed approach, by restricting  the acceptable justifications and 
explanations to commitments, it means that the proof terms $t$ in the 
$\mathcal{JEL}$ language represent commitments.


\begin{table}[t]
\begin{center}
\begin{tabular}{|l|l|}
\hline
  Expression & Informal Semantics\\
\hline
 $C(a,b,p,q):_i F$ & $C$ is probative evidence for $F$ for agent $i$.\\
 $\neg C(a,b,p,q):_i F$ & $C$ is not probative justification for $F$ for agent $i$.\\
 $(\neg C(a,b,p,q)):_i F$& The absence of $C$ is a justification for $F$ for agent $i$.\\
 $C(a,b,p,q):_i\neg F$ & $C(a,b,p,q)$ is evidence for $\neg F$ for the agent $i$.\\
\hline
\end{tabular}
\end{center}
\caption{Commitment based justification in a multi-agent system.}
\label{tab:expressivity2}
\end{table}

The classical definition of a conditional commitment states that
the debtor $x$ promises to creditor $y$ to bring about a particular
formula $P$ under the condition $Q$, encapsulated as $C(x,y,Q,P)$.
In multi-agent systems, a justification accepted as probative evidence
for an agent may not meet the standard of proof for another agent, which
rejects it.
To model this, we link the justification and explanatory operators to the agent
accepting the evidence.
Thus, the construction $C(a,b,p,q):_i F$ says that commitment $C$ is a
probative justification for agent $i$ regarding the sentence $F$ 
(see table~\ref{tab:expressivity2}).
A commitment may be preferred to another one by the agent $i$ when choosing 
an explanation, formalized as $(C\succ C)\triangleleft_i F$.
If one of the terms in the commitment is not constrained in any aspect, ``do not care'' sign ``$\_$'' is used. 
Examples~\ref{ex:da}, \ref{ex:pi}, and~\ref{ex:ej} illustrate how the 
$\mathcal{JEL}$ formalism is enacted in a commitment-based multi-agent setting.

\begin{example}[Distributed Application]
Assume the commitment $C(a,b,\_, F)$ is agent's $b$ justification for $F$ and 
the conditional commitment $C(b,c,F,G)$ represents a justification for
agent $c$ regarding the formula $F\rightarrow G$.
According to axiom $A_2$, the application operator builds a justification for $G$ of the form
$C(b,c,F,G) \cdot C(a,b,\_,F):_b G$.
Notice that the commitment aggregation is based on the view of agent $b$.
$$C(b,c,F,G):c F\rightarrow G \rightarrow (C(a,b,\_, F):_b \rightarrow C(b,c,F,G) \cdot C(a,b,\_,F):_b G)$$
\label{ex:da}
\end{example}
Not being omniscient, agents are not aware of all the consequences from their 
commitments. 
When used by the same agent, the proof checker operator acts as a positive 
introspection function, helping agents to be aware of the justification of 
their own commitments.
Given a commitment $C$ justifying the formula $F$, the agent $i$ can inspect 
its own commitment store to further justify the formula $C:_i F$ by enacting 
the application of the proof checker operator on itself: $!C:_i C:_i F$.
Consequently, given the commitment $C$ an agent can construct a justification 
for it based on other commitments.

\begin{example}[Positive Introspection]
Because agent $a$ has promised agent $b$ to deliver the $item$, given by $C(a,b,\top,item)$,
this is the justification of agent $a$ regarding the commitment $C(b,c,parts,pay)$ with agent
$c$, in which agent $b$ has to pay the components provided by
agent $c$.
Given the right associativity of the justificatory operator $:$, parenthesis are introduced only for clarity:  $$C(a,b,\top,item):_a [C(b,c,parts,pay):_a Item]$$.
\label{ex:pi}
\end{example}

\begin{example}[Explaining a Justification]
Although a hotel specifies in the commitment that check-in is at time 14,
usually there is no problem if the guest arrives at 12.
The situation in which the guest is not served, is often not justified
by personnel with the commitment itself (the active contractual relation)
but by other valid justifiers such as: "no clean rooms at the moment".
The customer can explain the situation to himself by the reduced number of
staff, which can be further explained by the economical policy of the
hotel, with the following chain of reasoning.
$$economicalPolicy\triangleleft_{guest} reducedStaff
\triangleleft_{guest} (\neg CleanedRooms) :_{hotel}\neg CheckIn$$
Justifying the impossibility to check-in by the existing contractual
commitment leads to a different chain of reasoning.
$$ increaseProfit \triangleleft_{guest} C(hotel,guest, pay, checkIn14):_{hotel}\neg CheckIn$$
\label{ex:ej}
\end{example}

Given the recurrent nature of business interactions, in common situations
business agents convey the same justificatory and explanatory schemes.
The following section formalizes two justificatory and two explanatory
patterns.

\section{Justificatory and Explanatory Business Patterns}
\label{sec:patterns}
An explanation involves discovering the meaning of an event in a particular
context, such that an $explanandum$ is explained by a coherent set of
$explanans$. 
The explanation aims to understanding the explanandum by indicating what
causes it. An explanation answers to $why$ questions or to contrastive 
\textit{why P rather than X} questions when the opponent requests 
an explanation for proponent preferences.


The theory of justification advocates the idea that justification is a
normative activity, where a concept is defined as normative if it depends
on norms.
The thing that justifies a proposition is called $justifier$.
Justifiers act as a vehicle between beliefs and knowledge, as the
definition of knowledge as justified true belief suggests.
Supplementary to explanans, a justifier should be legitimate by
objective factors such as social structures, normative frameworks, or
abstract rationality.


To synthesize the differences, a justification acts in a more normative
framework, whilst an explanation works in a social context.
An explanation implies the existence of an audience which understands the
claim and its explanans, whilst justification is in relation only with an
objective world of true beliefs.
The above observations are used to differentiate between two classes of business dialogs: justification patterns and explanatory patterns. 

\subsection{Justification Patterns}

Commitment-based justification patterns are constructed in terms of the
intention of the agent which provides the justification, the context in
which they are usually used, the formal representation of the pattern, within
specific cases.
An example is illustrated by informal text and its representation in
$\mathcal{JEL}$.
All commitment-based justification patterns of the general form
$C(a,b,P,Q):_c F$ may be questioned by a common set of critical
questions.
\begin{itemize}
 \item $CQ_1$: If $c\neq b$, which is the relationship between the agent
$c$ and $b$?
\item $CQ_2$: If $F\neq P$, which is the link between the
formulas $F$ and $P$?
\item $CQ_3$: Is there a stronger commitment which does not justify $F$?
\end{itemize}
Specific attacking options for each justification pattern are added at the
end of the pattern.

\subsubsection{Gratuitous Promise.}
In a gratuitous promise the debtor $x$ promises the creditor $y$ to bring
about $P$, without requesting anything: $C(x,y, \top,P)$.
This may serve as enough justification for agent $y$ for the achievement
of $P$, given by $C(x,y, \top,P):_y P$ (figure~\ref{fig:gp}).
The debtor is the one affected by the violation of the commitment.
In case the agent enacting the gratuitous promise is not the creditor, one
continuation path regards the relationship between the proponent of the
justification and the creditor.
In the example from figure~\ref{fig:gp}, in case $CQ_1$ is risen, the
relation between father and son can be legally justified.

\begin{figure}[t]
\begin{footnotesize}
\begin{schema}{Justification\ from\ gratuitous\ promise\doteqdot GP}
\begin{tabular}{ll}
$Intent$:& \textit{Stressing out that no one has forced the debtor to
commit.}\\
$Context$:& \textit{The creditor or other agent perform actions based on
the promise.}\\
$Pattern$:& \textit{$C(a,b,\top,P):_c F$}\\
$Variants$:& \textit{b=c, a=c, F=P}\\
\end{tabular}
   \where
\begin{tabular}{ll}
$Example$: &\textit{Grandfather promises to his nephew to pay for a trip.} \\
        & \textit{Based on it, the father buys a new bag for his son.}\\
$\mathcal{JEL}$: & \textit{$C(grandfather,nephew,\top,PayTrip):_{father}
NewBag$}\\
\end{tabular}
\end{schema}
\end{footnotesize}
\caption{Commitment-based justificatory promise.}
\label{fig:gp}
\end{figure}

For the $CQ_2$ attack, the justification stands only if there is an
evident connection between the promise $P$ and the action $F$.
Note that the particular variant $C(a,b,\top,P):_b P$ avoids the above two
attacks.
Assuming that in sequent step, the nephew informed that he has no bag for
the trip, 
the grandfather reacts by assuring his nephew that he will pay both for
the trip and buy a new bag. 
From the father's perspective, the newly stronger commitment acts as a
justification for the opposite conclusion $\neg NewBag$: 
$$C(grandfather,nephew,\top,PayTrip \wedge NewBag):_{father} \neg NewBag$$

The promise made by the creditor can justify future actions of the
creditor itself.
For instance the grandfather starts saving money due to its gratuitous
promise: $C(grandfather,nephew,\top,PayTrip):_{grandfather} SaveMoney$.

\begin{figure}[t]
\begin{footnotesize}
\begin{schema}{Justification\ from\ request\ \doteqdot R}
\begin{tabular}{ll}
$Intent:$ &  \textit{Justifying actions based on the directive conveyed by
a }\\
& \textit{normative empowered agent.}\\
$Context:$ &   \textit{The creditor or other agent act on request from
other agent.}\\
$Pattern:$ & \textit{$C(a,b,P,\top):_c F$}\\
$Variants:$ & \textit{b=c, a=c, F=P, $\neg P$}\\
\end{tabular}
   \where
\begin{tabular}{ll}
$Example:$ & \textit{During driving lessons, the instructor requests the
student to stop }\\
& \textit{the car. Consequently, the next student's action is to signal
right.} \\
$\mathcal{JEL}$: & \textit{$C(instructor,student,StopCar,\top):_{student}
SignalRight$}\\
$CQ_4:$ &\textit{Is the request legitimate?}\\
\end{tabular}
\end{schema}
\end{footnotesize}
\caption{Commitment-based justificatory patterns.}
\label{fig:jp}
\end{figure}

\subsubsection{Directives.} In a \textit{fact request} the debtor does not
promise anything, it only requests the precondition $q$ to be satisfied,
given by $C(a,b,q,\top)$ (figure~\ref{fig:jp}).
The additional critical question $CQ_4$ regards the normative rightness of
the directive.
For instance, there is no obligation for the agent $b$ to meet the request
or, if there is indeed a power relationship between the creditor and the
debtor, it may not have jurisdiction in the context of the sentence $q$
requested.
In the example of figure~\ref{fig:jp}, the requested action of stopping
the car is legitimate by the relationship between the instructor and the
student.
If the requested act is negated, the commitment represents a $taboo$ or an
interdiction.

\begin{example}[Justification from Interdiction]
"I can not sell you cigarettes because the law interdicts to sell them to
the minors" will be formalized in $\mathcal{JEL}$ as 
$$C(na,\_, \neg SellCigarMinor,\top):_{me}\neg
C(me, you, Pay, SellCigar):_{me} \neg Sell$$
 Here the law is personalized by the normative agent $na$.
According to contract law, when exposing an item for selling, an open
offer is created.
The seller commits to sell the item in case of acceptance, which usually
occurs by payment: $C(seller, buyer, Pay, Sell)$.
The exact representation captures this when the agent $me$ justifies his
refuse to sell, because he is not committed to do so, which is further
justified by the normative interdiction to sell cigarettes to the minors.
\end{example}

The two justificatory patterns introduced above are not exhaustive in terms of basic justificatory schemes.
However, by combining such basic patterns one can increase the expressivity of the justifiers. 
Firstly, by composing a gratuitous promise with a fact $Q$ treated as a condition, a \textit{unilateral contract} $UL$ is obtained: $C(a,b,\top,P)\circ_q Q = C(a,b,Q,P)$. 
The same $UL$ pattern can also be obtained by composing a request with a fact $P$ treated as a promise: $C(a,b,Q,\top) \circ_p P = C(a,b,Q,P)$.  

Secondly, when the term used for composition is a commitment itself 
a higher order commitment~\cite{LetiaG06} is constructed.
In this line, \textit{justification from bilateral contract} $BC$ is formalized
as $C(a,b,C(b,a,\top, pay),deliver)$. 
Here, both parties make promises: the agent $a$ commits to deliver the item if 
the agent $b$ promises to pay. 
The pattern is a composition between a unilateral contract and a justification 
from request pattern. 
The composition is applied on the requested action which represents 
a commitment itself: $C(a,b,Q,P) \circ_q C(b,a,\top,Q_1)$. 
For the given example, the justifier is constructed as follows: 
$BC=C(a,b,_,deliver)\circ_q C(b,a,\top, pay)$.
The justification from \textit{promise to commit pattern} is a composition of 
two gratuitous promises, applied on the fourth term of the commitment: 
$C(a,b,\top,\_) \circ_p C(a,c,\top,pay)=C(a,b,\top,C(a,c,\top,deliver))$, 
where agent $a$ promises agent $b$ that he will commit to agent $c$ 
to deliver the item.

\subsection{Explanation Patterns}

\begin{figure}[t]
\begin{footnotesize}
\begin{schema}{Explanation\ from\ cognitive\_consistency\ \doteqdot CC}
\begin{tabular}{ll}
  $Intent:$ & \textit{Explaining actions based on the goals that an agent
is following.}\\
   $Context:$ & \textit{The debtor commits itself to achieve a particular
sentence.}\\
   $Pattern:$ & \textit{$C(a,a,Q,P)\triangleleft_c F$}\\
   $Variant:$ &\textit{a=c, $Q=\top$}\\
\end{tabular}
   \where
\begin{tabular}{ll}
$Example:$ & \textit{Tom cannot join the party because he wants to learn
for the exam.}\\
$\mathcal{JEL}$: & \textit{$C(tom,tom,\top,Learn)\triangleleft_{jim} \neg JoinParty$}\\
$CQ_4$: & \textit{Is the debtor aware of his commitments?}\\
\end{tabular}
\end{schema}
\end{footnotesize}
\caption{Explanation for preferred commitment.}
\label{fig:pc}
\end{figure}

Explanation patterns are not rooted in an objective normative frame, having
a subjective component.
Explanatory schemes can be viewed as providing subjective reasons, with a
more flexible relation between explainers and what is explained.

\subsubsection{Public goals.}
Consider the situation in which the agent commits itself to bring
about $r$: $C(a,a,\top,r)$.
It is on the edge between social semantics and mentalistic semantics of
communicative agents (figure~\ref{fig:pc}).
The commitment belongs to the social semantics because it is public, and
points toward the mentalistic approach because it represents a goal of the agent.
Assuming sincere agents, the attack option $CQ_4$ questions the
possibility that the agent may not be aware of all its commitments.

\begin{figure}[t]
\begin{footnotesize}
\begin{schema}{Explanation\ from\ preferred\_commitment\ \doteqdot PC}
\begin{tabular}{ll}
$Intent:$ &    \textit{Explaining choice between two commitments.}\\
$Context:$&    \textit{The debtor is committed with different strengths to
creditors.}\\
$Pattern:$ &    \textit{$C(a,b,P,Q)\succ C(a,c,P',Q')\triangleleft_d F $}\\
$Variants:$ &   \textit{a=d, b=d, c=d}\\
\end{tabular}
   \where
\begin{tabular}{ll}
$Example:$ & \textit{The agent $a$ promised his boss to attend at a
late meeting. He also}\\
& \textit{promised his wife to take the child from the school if he has
time.}\\
& \textit{Aware of the constraint, the wife decides to go to the school
herself.}\\
$\mathcal{JEL}$: & \textit{$C(a,boss,\top,Meeting)\succ C(a, wife,Time,
TakeChild)\triangleleft_{wife} GoSchool$}\\
$CQ_4:$ &\textit{Is it not possible to achieve the both commitments?}\\
$CQ_5:$ & \textit{Is the preference relation explained by the cognitive
consistency}\\
& \textit{property of the agent $d$?}\\
\end{tabular}
\end{schema}
\end{footnotesize}
\caption{Commitment-based explanatory preference.}
\label{fig:ep}
\end{figure}

\subsubsection{Explanation from preference.}
The preference relation usually has a strong subjective component, making
it a candidate for explanatory arguments rather than justificatory ones.
In our example (figure~\ref{fig:ep}), $C(a,boss,\top,Meeting)$ 
is a gratuitous commitment, whilst $C(a, wife,Time, TakeChild)$ 
is a unilateral contract. 
The gratuitous promise representing a stronger promise compared to a unilateral 
contract, the preference relation can be deduced based on the strength of each 
type of commitment.  

\section{Argumentation Framework}
\label{sec:agents}

\subsection{Arguing with Justification and Explanation}

\begin{definition}
An argument is a pair $\langle t,F \rangle$ where $t$ is the chain 
of justifiers or
explanans and $F$ the conclusion such that $t$ justifies or explains
$H$ and $t$ is minimal.
\end{definition}

We distinguish between \textit{explanatory arguments}  and
\textit{justificatary arguments}.

\begin{definition}
A justificatary argument is supported by justifiers only. An explicatary
argument contains at least one supporting explanandum.
\end{definition}
A fact can be supported at the same time by several explanatory or
justificatory arguments.

\begin{definition}[Conflict among arguments]
In the case of an undercutting argument, its conclusion attacks one justifier or explanation
in the support of another argument.
In the case of rebuttals, the justified formulas contradict each other directly.
\end{definition}

There are two types of interrogative requests: request for explanation and
request for justification.

\begin{definition}[Request for justification]
Agent $a$ requests agent $b$ to provide its justification why t is agent its
justification for $F$: $C(a,b,!t:_b t:_i F,\top)$.
\end{definition}

The general case reflects the situation when the agents $a$, $b$ and $i$
are different.
\begin{example}[Request for justification]
Consider that the $judge$ agent requests the $lawyer$ agent to legally
support the belief of the victim $v$ that $selfDefense$ is a justification
for $useGun$, expressed as: 
$$C(judge, lawyer, !selfDefense:_{lawyer} selfDefense:_{v} UseGun,\top)$$
Note that the justification $!selfDefense$ is requested to be constructed 
from the perspective of the lawyer agent.
The expressivity of the language allows to ask the lawyer to present the
victim's justification regarding the sentence 
$selfDefense:_{v} UseGun$, given by $C(judge, lawyer, !selfDefense:_{v}
selfDefense:_{v} UseGun,\top)$. 
\end{example}

When $b=i$, the agent $b$ is requested to justify itself.
When $a=b$, the agent $a$ requests itself to identify or construct 
a justification supporting why $t$ is accepted by the agent $i$ as valid justifier for $F$.
Not being omniscient, in $\mathcal{JEL}$, the agents are not
assumed to be aware of all the justifications that can be built from their
knowledge base.
Only an explicit constructed justification is considered, which makes
sense for the agent to interrogate its own knowledge base to identify a valid
justification.
Similarly, when $a=b=i$, the same agent is applying itself to the
task of justifying its own sentences.

\begin{definition}[Request for explanation]
Agent $a$ requests agent $b$ to explain why $t$ is the justification of $i$
for $F$: $C(a,b,!t \triangleleft_j t:_i F,\top)$.
Similarly, agent $a$ requests agent $b$ to further explain why $t$ is
the explanation of agent $i$ for 
$F$: $C(a,b,!t \triangleleft_j t\triangleleft_i F,\top)$.
\end{definition}

Notice that one cannot justify an explanation, but an explanation can be
requested both for a justifier and for another explanation.
The combination of the above explicatory/justificatory patterns in
dialogs helps the parties to understand normative-based decisions.

\subsubsection{Strength of Justification.}
The strength of a justification depends on the commitment used as evidence
and the formula needed to be justified.
The stronger the commitment used as a justification term, the stronger the
justification.
\begin{proposition}
The commitment $C(a,b,q,p)$ is stronger than $C(a,b,q',p')$ if the debtor
promises more and requests less.
\end{proposition}

\begin{example}[Strength of justification]
"The supplier $s$ commits to deliver more items faster to the retailer
$r$" is stronger than "The supplier commits to deliver the items earlier
if he receives the payment earlier": 
$$C(s, r,\top, MoreItems \wedge FasterDeliv) \gtrdot C(s,r,PayEarlier, FasterDeliv)$$
\end{example}

In the particular case of no promised action ($p=\top$), the same
semantics works for defining stronger requests.
Thus, the request pattern $C(a,b,p,\top)$ is stronger or more specific
than $C(a,b,p\vee q,\top)$.
This can be used during negotiation dialogs, where the less stronger request
allows more options for the requester and introduces higher flexibility in 
the system.
\begin{example}
 The request to pay either by credit card or by wire transfer is
stronger to the wire transfer option, formalized as: 
$C(bank,a,card \vee wired, \top) \gtrdot C(bank,a,wired,\top)$.
\end{example}

\begin{proposition}
A fact $p$ is stronger than the commitment created to bring about that $p$, given by $p \gtrdot C(\_,\_,\_,p)$.
\end{proposition}

\begin{example}
"I commit to deliver the item after you confirm the order" is stronger then
"I commit to deliver the item if you have promised me to confirm the
order": 
$C(me, you,Confirm,Deliver) \gtrdot C(me, you, C(me, you, \top,Confirm),Deliver)$.
\end{example}

A \textit{preference relation} $succ$ on commitments can be defined based on the strength of the commitment.
\begin{proposition}
The creditor prefers stronger commitments, whilst the debtor prefers
weaker commitments when explaining the formula $F$. 
If $C(a,b,P,Q) \gtrdot C(a,b,P',Q')$ then we have the following.
$$ C(a,b,P,Q) \succ C(a,b,P',Q') \vartriangleright_a F\  \text{and} \ 
C(a,b,P',Q') \succ C(a,b,P,Q) \vartriangleright_b F$$
\end{proposition}

\subsubsection{Conflict Resolution.}
In $\mathcal{JEL}$ all arguments are considered to have
justifications or explanations.
When deciding between two conflicting formula, the stronger justified
formula will be preferred by the debtor, because the stronger commitment
provides more guarantees for the promised action.
If $C(a,b,p,q):_b F$ and $C(a,b,\top, q \wedge r):_b\neg F$, the $F$
formula would be accepted.
In case the justifications have the same strength,
$\mathcal{JEL}$ forces the agents to provide further justifications on request.

\begin{definition}[Preference-based conflict resolution]
If $F$ attacks $F'$ and $F'$ attacks $F$ and
$c_n \vartriangleright c_{n-1}\vartriangleright ...c_i\vartriangleright c_j\vartriangleright ....\vartriangleright c_1 \vartriangleright F$ and
$c'_n\vartriangleright c'_{n-1}\vartriangleright...c'_{i+1}\vartriangleright c'_i\vartriangleright....:c'_1\vartriangleright F'$, $F$ is preferred to $F'$ if
there is a chain of explanans $c_1\vartriangleright...c_k$ so that
$equalStrength(c_k, c'_k)$ $\forall k \in [1,i]$ and $c_{i+1} \succ
c'_{i+1}$.
\end{definition}

\subsection{Argumentative Agents}
Combining commitments with justification and explanation logic provides flexibility for defining several types of argumentative agents. 

Firstly, regarding the logical framework, in the simplest approach, three types of agents can be formalized: 
The less demanding agent requires only explanations in order to accept a formula, given by the axiom $t\triangleleft_i F \rightarrow F$.
A rigorous agent will accept only normative justified formulas: $t:_i F \rightarrow F$. 
The most demanding one requests both explanation and justification before accepting a sentence: $t\triangleleft_i F \wedge s:_i F \rightarrow F$.
Aggregation of these components can lead for instance to a $caution$ agent, which accepts a formula $F$ if it has a valid justification and no valid explanation supporting the opposite conclusion $\neg F$, formalized as: 
$$t:_i F \wedge (\neg s) \triangleleft_i \neg F\rightarrow F$$

In a more elaborate agent system, the existing theories of justification (foundationalism, infinitism, internalism, externalism~\cite{sep-reasons-just-vs-expl}) and explanation (causal, teleological)~\cite{Little91}) can be exploited to define the corresponding agent type. 
Each theory requires different amount and type of evidence before a formula can be considered justified or explained. 
For a $foundationalist$ agent, the existence of a basic justificatory pattern would be enough for accepting the supported formula. 
Starting from the infinitism theory, a \textit{n-type credulous} agent accepts a formula if its justification chain has at least length $n$.
An $internalist$ agent should be able to justify a sentence only through its own commitments, whilst for an $externalist$ agent, third party commitments can be used as justifiers. 

Secondly, regarding the commitment patterns, agents can convey or accept as valid justifiers only commitments meeting a strength threshold.  
By composing the basic gratuitous promise and request justificatory patterns, one obtains higher order patterns having different degrees of strength. 
Table~\ref{tab:risk} illustrates possible composed patterns that may justify that the payment will be made. 

By promising to deliver an item and only requesting a promise for payment, the pattern $GP+(R\circ_q GP)$ is the weakest one, when justifying the formula $Pay$. 
Only an agent having a low justification standard as \textit{scintilla of justification} would accept this pattern as valid justification. 
A little bit stronger justification is given by the pattern $GP+R$ where a promise for delivering and a request for paying do exist.  
The agent $a$ should have the \textit{reasonable justification} proof standard to accept this justifier.
In the bilateral contract $UC\circ_q GP$ the promise to pay representing a precondition for delivering, gives \textit{preponderence of justification} for the agent $a$ to consider it as a valid justifier. 
The pattern $UC$ meets the \textit{convincing justification} standard of proof of the agent $a$, whilst the most attractive pattern for the agent $a$ is $UC\circ_p GP$, where it promises to deliver the item only after the payment is finalized. 
This justifier should meet the most skeptical standard of proof of an agent, which is \textit{behind any reasonable doubt}.
 
\begin{table}
\begin{scriptsize}
\begin{center}
\begin{tabular}{|l|l|l|}
 \hline
  \textbf{Pattern} & \textbf{Justifier} & \textbf{Meaning} \\
  \hline
$GP+(R\circ_q GP)$ & $C(a,b,\top,deliver) +$   & $a$ commits to deliver the item and\\
 & $C(a,b,C(b,a,\top,pay),\top):_aPay$ & requests $b$ to commit to pay for it.\\
\hline
  $GP+R$   & $C(a,b,\top,deliver)+$ & $a$ commits to deliver the item\\
  & $C(a,b,pay,\top):_aPay$&and requests $b$ to pay for it.\\
\hline
  $UC\circ_q GP$& $C(a,b,C(b,a,\top,pay),deliver):_a Pay$ & $a$ commits to deliver the item\\
 &  & if $b$ commits $a$ to pay for it.\\
\hline
 $UC$ & $C(a, b, pay,deliver):_aPay$ & $a$ commits to deliver the item\\
    & & in case $b$ pays for it.\\
\hline
 $UC\circ_p GP$ & $C(a,b,pay,C(a, b,\top, deliver)):_a Pay$ & $a$ will commit to deliver the item if $b$ pays.\\
\hline
\end{tabular}
\end{center}
\end{scriptsize}
\caption{Composed patterns with varying degrees of strength.}
\label{tab:risk}
\end{table} 

Thirdly, with respect to the explanatory component, the cognitive consistent pattern is used by $sincere$ agents. 
Such an agent is committed to itself to bring about the items that he promised to perform.
It comes with several flavors:
An $wholehearted$ agent $x$ would try  to bring about $p$ if he promised
to do so, no matter who its partner is and no matter if its partner has
already performed or not the requested action:
$C(x,x,C(x,\_,\_,p), p)$.
Some agents can manifest sincerity only to some partner $y$:
$C(x,x,C(x,y,\_,p), p)$. 
Here, if the agent $x$ is committed to the particular agent $y$
than he is committed to itself to bring about $p$.
Re-assurance can be provided to its partner: ``If I have promised you to bring
about $p$ I will.'', given by $C(me,you,C(me,you,\_,p), p)$ or to a supervisor: 
``I am committed to my boss to keep
my promises to all agents which do not have to perform something in exchange.'', formalized as $C(me,boss,C(me,\_,\top,p), p)$. 

Quite differently, for a $diffident$ agent $x$, even he has promised to
bring about $p$ this is not conclusive for itself that $P$ will hold,
given by  $\neg C(x,y,\top,P):_x P$.
Depending on the justification provided for this sentence, it can be:
i) the agent simply does not trust its capabilities to accomplish the task;
ii) the agent is aware of some stronger commitment that may block the execution of $P$; or
iii) he is not committed to itself to satisfy its own promises and it is aware of this.

\section{Trip Booking Scenario}
\label{sec:scenario}

\begin{figure}
\begin{center}
\begin{tikzpicture}[scale=.9, transform shape]
\tikzstyle{arg} = [circle, fill=gray!30, ]
\tikzstyle{arg1} = [shade=ball, circle, draw]
\tikzstyle{argbox} = [fill=gray!0]
\tikzstyle{role} = [above, text width=5em,text centered]

\node[arg1] (s2) at (-2.5,-0.5) {$s_1$};
\node[arg1] (s1) at (-2.5,-1.5) {$s_2$};
\node[arg1] (t) at (1,-1) {$t$};
\node[arg1] (c) at (3,-1) {$a$};

\draw [draw,->] (s1) -- (t);
\draw [draw,->] (s2) -- (t);
\draw [draw,->] (t) -- (c);

\node[arg1] (na) at (2,1) {$na$};
\draw [draw,->] (na) -- (t);
\draw [draw,->] (na) -- (c);

\node[argbox,text width=20em] (9) at (8.4,0) {Commitment Store:\\
$C(t,a,pay_3,trip)$\\
$C(t,t,flight\wedge acc, trip)$\\
$C(s_2, t, pay_1,acc)$\\
$C(s_1,t,C(t,s_1,\top,pay_2),flight)$\\
$C(na, t,\neg trip  \wedge pay_3, C(t,\_,\top, pay_4))$\\
$C(na, \_, \neg EUcitizen, visa)$\\
$C(na,\_,swiss,\neg visa)$};

\end{tikzpicture}
\end{center}
\caption{Commitment store in the running scenario.}
\label{fig:scenario}

\end{figure}
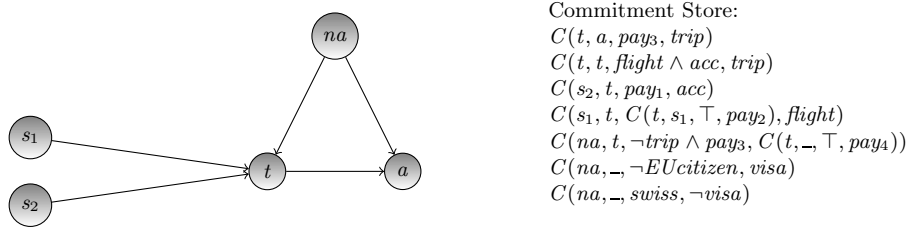
Consider the scenario in which the agent $a$ wants to book a trip to
Valencia from the tourism agency $c$.
The company has contracts for flights with the air company $s_1$ and for
accommodation with the business entity $s_2$ (figure~\ref{fig:scenario}).
The agent $t$ promises agent $a$ to deliver the trip after he pays for it, represented by the $UC$ pattern $C(t,a,pay_3,trip)$.
The capabilities of the agent $t$ are represented by the $CC$ explanatory pattern $C(t,t,flight\wedge acc, trip)$ in which, if he manages to book the flight and accommodation, he can deliver the requested trip for his client. 
The tourism agency is aware of the unilateral contract from the accommodation
company $s_2$, given by $C(s_2, t, pay_1,acc)$, and also of the bilateral contract issued by $s_1$: $C(s_1,t,C(t,s_1,\top,pay_2),flight)$.
Here, the $s_1$ agent will provide plane tickets if the partner agent $t$
promises to pay the amount $pay_2$.

Given the location of the agent $t$ within the European Community,
it is doing business under the jurisdiction of the corresponding normative agent $na$.
Following the contract law, the normative agent $na$ requests to all its
business entities under its umbrella that an open offer accepted by a client
should be honored, otherwise the breaching agent should return the money
and also pay penalties.
The composition between a unilateral contract and a gratuitous promise $UC\circ_p GP$ is used to model this: 
$$C(na, t,\neg trip  \wedge pay_3, C(t,client,\top, pay_4))$$ 
Here, if the client pays the amount $pay_3$ and he does not receive the trip, the tourism company is committed to pay the amount $pay_4=pay_3+penalty$ to the client.    

Assume that the tourism agency is an \textit{internalist 1-type credulous} agent with a \textit{convincing justification} standard of proof. 
The client $a$ is \textit{externalist, cautious} agent having the \textit{preponderance of evidence} justification standard. 

The agent $a$ commences the dialog by requesting justification $j$ guarantying the trip in case he pays, given by the $R$ pattern:

$$C(a,t,j:_t Trip,\top)$$
Being an internalist agent, the tourism company $t$ can guarantee the trip based only on its contractual clauses with the suppliers $s_1$ and $s_2$:
$$[C(s_2, t, pay_1,acc) + C(s_1,t,C(t,s_1,\top,pay_2),flight)]:_t Trip$$
We note the compound justifier above with $J$, such that $J:_t Trip$.
The first term in $J$ is a UC, whilst the second commitment represents a BC. 
Both terms satisfying the proof standard of agent $a$, the compound justifier $J$ will meet the justification standard of agent $a$. 
Being an externalist agent, the client can accept the justification of the agent $t$ as valid:
$$[C(s_2, t, pay_1,acc) + C(s_1,t,C(t,s_1,\top,pay_2),flight)]:_a Trip$$
Note that in this line of justification, the agent $t$ releases some private information like
the values $pay_1$ and $pay_2$.

Being a cautious agent, the client should check if there are explanations supporting the opposite conclusion $\neg Trip$. 
Firstly, it requests explanans why $J$ offers enough justification for the formula $Trip$:
$$C(a,t,!J\triangleleft_t J:_t Trip,\top)$$
Using the $CC$ explanation pattern, the tourism agency explains based on its business practice that plane tickets and accommodation are enough to provide the requested trip:
$$ C(t,t,acc\wedge flight, trip) \triangleleft_t J:_t Trip$$

Consider that the agent $a$ is aware of the regulation requesting visa for non EU
citizens, formalized as $C(na, \_, \neg EUcitizen, visa)$. 
This is the main concern of the agent $a$ for rebutting the $Trip$ formula.
$$C(na, \_, \neg EUcitizen, visa):_a J:_a \neg Trip$$
We note the above justification chain with $J'$ such that $J':_a \neg Trip$.  
At this moment the tourism agency may request explanations to agent $a$ regarding his concern:
$$C(t,a,!J'\triangleleft_a J':_a \neg Trip, \top)$$

The agent $a$ can provide the explanation that he is not an EU citizen.
This helps agent $t$ to figure out the case, but being a 1-type cautious agent it needs a justification to accept the formula $\neg EUcitizen$. 
Such a justification is requested through:
$$C(t,a,J'':_a\neg EUcitizen, \top)$$
One option would be to object the request by activating the critical question $CQ_4$ of the $R$ pattern, considering that the agent $t$ is not legitimate to ask for such justifications. 
The other option is to provide an ID card or passport acting as a justification: $swissPassport:_a \neg EUcitizen$. 
A fact being stronger that a commitment according to proposition 2, it satisfies the preponderance of justification standard of the $t$ agent.
Note that it may represent a constant in the justificatory logic framework.

Knowing that Swiss citizens do not need a visa for traveling in Europe, the commitment $C(na,\_,swiss,\neg visa)$ undercuts the $a's$ justifier supporting $\neg Trip$ formula:
$$ C(na,\_,swiss,\neg visa):_t \neg J':_a \neg Trip$$
We note the above justifier with $J'''$ such that $J''':_t\neg J':_a \neg Trip$. 
$J'''$ commitment refers to the agent $t$ too, which is under the normative framework of the agent $na$. Therefore, it can be used by the internalist agent $t$ to constructs justifications.
Being cautious, the agent $a$ needs a supplementary explanation, requested with: 
$$C(a,t, !J'''\triangleleft_t J''':_t \neg J':_a \neg Trip,\top)$$ 
The EU preferring to encourage traveling across Europe from safe countries, rather then imposing unnecessary security constraints, explains why a visa is not required:
$$[C(na,na,\top,encourageTravel) \succ C(na,\_,\neg EUcitizen, visa)] \triangleleft_t J''':_t\neg J':_a \neg Trip$$

Having both a justification and an explanation, the agent $a$ accepts that his visa concerns are defeated.
Consequently, given the accepted valid justification $J$ for the trip, and no valid explanation for $\neg Trip$, the agent accepts the $Trip$ formula.

\section{Discussion and Related Work}

\paragraph{Justification Logic and Argumentation.}
In the proposed approach we apply the justification logic mechanism to an argumentation context.
The connection between explanation, justification, and argumentation is best evidenced in scenarios where
understanding is the focus (eg. learning). 
One reason is that the constructivist approach needed in understanding is employed by justification logic.
Justification Logic treats justification as the primary object, whilst the claim is secondary.
Quite the opposite, nonjustificational criticism works towards attacking claims themselves, which is closer to classical approach in the argumentation theory.
The proof checker operator of JL promotes reflective communication or deliberation on justificatory and explanatory knowledge.

The constants in JL can be mapped to \textit{prima facie} arguments, which
do not require any sort of justification, whilst standard arguments should
be supported by justification chains.
The framework of $\mathcal{JEL}$ assures that all arguments are
supported through a chain starting from \textit{prima facie} arguments
only.

The proposed framework applies the JL formalism to a multi-agent system. 
Differently from existing approaches~\cite{Yavorskaya08,LetiaG11}, we introduce the explanatory operator and force the justifiers to be social commitments. 
Particular to the classical approaches in JL~\cite{Fit09TL}, the explanatory operator introduces non-monotonic reasoning in the JL framework. 
One of the advantages that commitments bring in this justification logic landscape, is that there are more possibilities to define communicative acts between agents: like requests or promises between conveying agent $a$ and receiving agent $b$.

\paragraph{Argumentation Dialogs.}
In the dialog typology of Walton and Krabbe~\cite{Walton95},
information seeking and inquiry dialogs implies a search for a true
answer to some factual question~\cite{BlackH09}.
In this respect, justification and explanation are subtypes of such
dialogs.
In the Habermas discourse theory~\cite{Habermas81}, the classification includes explicative,
theoretical, and moral discourses, where explicative discourses aim at
increasing comprehensibility, whilst theoretical discourses try to
discover the truth.

A first step towards integrated argumentation reasoning with explanatory
reasoning is made in~\cite{Bex10}, where the distinction between argumentation and
explanation does not come from the statical structure or mode of
reasoning (often abductive for explanation, mostly deductive in
argumentation), but rather from broader dialogical context.
Justifying reasons appear in two classes: epistemic reasons and practical
reasons, treated as arguments~\cite{AmgoudP09}.
Epistemic justifications (or theoretical reasons) are based only on
beliefs and are used to justify beliefs.
Practical justifications are constructed from beliefs and preferences and
used to justify options or actions.

Compared to argumentation schemes, which aim to capture domain
independent argumentation patterns, our proposal is more business oriented.
We try to construct basic argumentation blocks used in negotiation and
business process monitoring.
This represents a step towards developing an \textit{argumentation pattern
language} directly applied on business process modeling languages.

Based on the content of explanation, rule-based systems exploit four types
of explanations: (i) $trace$, which shows the line of reasoning supporting
the conclusion,  (ii) $justification$, describing the rationale behind
each reasoning step, (iii) $control$ or $strategic$ explanations, which
bears out the problem solving strategy or control behavior of the system,
and (iv) $terminological$, providing definitional information~\cite{Yetim08}.
Justification-type explanations seems to give rise to more
positive user perceptions of a knowledge-based system than trace and strategic explanations~\cite{Ye95}.  
Many knowledge-based systems address the issue of presenting user-adapted explanations~\cite{Yetim08}.

Explanation types depend on domain knowledge. Within the social domain, folk
psychological explanations are conveyed, in the physical domain both naive and sophisticated
theories, whilst in the religious domain explanatory standards prevail. 
In our case, the main goal of seeking explanation in
the business domain is to increase the ability to make predictions and to understand normative-based decisions.

\paragraph{Social Commitments.}
Explanation and justification are treated here as social constructs.
Commitments have been used within an argumentation framework
in~\cite{BentaharMMC04}, where argumentation relations as defend, attack, or justify are defined between
two commitments. In our case, the strength relation is inferred based on the commitment pattern, whilst the preference relation depends on the agents role in the commitment, as creditors or debtors. 
In both approaches, the agents should be able to justify their commitments.

The expressivity arising from combining commitments have been exploited
for representing business patterns~\cite{ChopraS11,Telang11TSC} or legal contracts~\cite{LetiaG06}.
The large amount of work on commitments deals with modeling business
patterns, which has been proved satisfactory for real life business
protocols.
Often, a dialog occurs in which the creditor requests for justifiers
or performance, whilst the debtor provides the requested justification or explanation in
order to re-activate the commitment or to find an alternative
solution.

Defining commitment-based protocols is easier due to the closed word assumptions when designing a protocol.
This is not the case in argumentation where agents may convey in an open
world different types of justifications and explanations.
That is why an argumentation process is needed on top of the commitments
in order to assure the flexibility encountered in real life business
interactions.

\section{Conclusions}
The main goal of our study has been to propose a technical instrumentation for handling both justificatory and explanatory arguments.
The first contribution regards the setting of some basis of exploiting in computational models of arguments the differences between justification and explanation, as already stressed out in the philosophy of science. 

As a second contribution, the $\mathcal{JEL}$ is developed to cover both justifiers and explanans, in the line of using logic in argumentation as envisaged by Gabbay~\cite{Gabbay10}. 
Introducing commitments enhances the capability of agents to reason over justifications and explanations of the other agents.

A third contribution regards the formalization of justificatory and explanatory commitment-based patterns. 
The individual actions are taking place within a framework of interdependent social commitments.
We consider agents as placed in various networks of commitment relations with other agents, where social influences formulate the justification behind agents decisions~\cite{Karunatillake05}. 
The main benefit here regards the flexibility to construct a large variety of higher order patterns. 

Using together $\mathcal{JEL}$ with commitments provides opportunities for defining several types of argumentative agents. 
Activating specific axioms in $\mathcal{JEL}$, rigorous or cautios agents can be formalized. 
The justification standard of each agent is defined based on the strength relation between compound commitments.   

As future work we will be investigating the role of critical questions 
in blocking justifications and explanations.

\section*{Acknowledgements}
We are grateful to the anonymous reviewers for their useful comments. 
Adrian Groza is supported by the Sectoral Operational Programme Human Resources Development 2007-2013 of the Romanian Ministry of 
Labour, Family and Social Protection through the Financial Agreement POSDRU/89/1.5/S/62557.

\bibliographystyle{splncs03}
\bibliography{argmas2011,jl}

\end{document}